\documentclass[preprint,12pt]{elsarticle}

\usepackage{amssymb}
\usepackage{amsmath}
\usepackage{algorithm}
\usepackage{amsmath}
\usepackage{amssymb}
\usepackage{xcolor}
\usepackage{booktabs}
\usepackage{multirow}
\usepackage{algpseudocode}
\usepackage[symbol]{footmisc}
\usepackage{appendix}
\usepackage{makecell}
\usepackage{hyperref} 
\usepackage{lineno}

\usepackage{zref-base}
\usepackage{zref-lastpage}
\makeatletter
\zref@newprop{linenumber}{\the\c@linenumber}
\zref@addprop{main}{linenumber}
\makeatother

\def\SD#1{{\color{black} {{#1}}}} 

\begin{document}

\begin{frontmatter}
\nolinenumbers

\title{4DLoG: Generative Modeling of Neurodegenerative Brain Anatomy with 4D Longitudinal Diffusion Model}

\author{Nivetha Jayakumar\footnote{These authors contribute equally to this work.}} 

\affiliation{organization={Department of Electrical and Computer Engineering},
            addressline={University of Virginia}, 
            city={Charlottesville},
            postcode={22903}, 
            state={VA},
            country={USA}}

\author{Swakshar Deb\textsuperscript{*}} 

\affiliation{organization={Department of Electrical and Computer Engineering},
            addressline={University of Virginia}, 
            city={Charlottesville},
            postcode={22903}, 
            state={VA},
            country={USA}}

\author{Bahram Jafrasteh} 

\affiliation{organization={Department of Radiology},
            addressline={Weill Cornell Medical College, Cornell University}, 
            city={Ithaca},
            postcode={14850}, 
            state={NY},
            country={USA}}            

\author{Qingyu Zhao} 

\affiliation{organization={Department of Radiology},
            addressline={Weill Cornell Medical College, Cornell University}, 
            city={Ithaca},
            postcode={14850}, 
            state={NY},
            country={USA}}

\author{Miaomiao Zhang} 

\affiliation{organization={Department of Electrical and Computer Engineering, Department of Computer Science},
            addressline={University of Virginia}, 
            city={Charlottesville},
            postcode={22903}, 
            state={VA},
            country={USA}}

\begin{abstract}
Modeling and predicting neurodegenerative disease progression from medical images remains a major challenge in medical AI, with significant implications for early diagnosis, disease monitoring, and treatment planning. However, most longitudinal neuroimaging datasets are temporally sparse, with substantial gaps and missing follow-up scans for individual subjects. This makes it difficult to learn and accurately capture the continuous anatomical changes associated with disease progression at the level of individual subjects. \textcolor{black}{To address this problem, we propose a novel model, named as {\em 4DLoG}, a full \underline{4D} (3D$\times$T) \underline{L}ongitudinal \underline{G}enerative framework that effectively models and synthesizes follow-up brain anatomy over time, conditioned on available clinical and demographic variables. In contrast to previous approaches, our 4DLoG features two main contributions. First, it introduces a full 4D generative diffusion framework that jointly models spatial and temporal dependencies across complete longitudinal sequences through dedicated spatiotemporal attention, with robust spatial patch extraction and temporal alignment. Second, it explicitly learns the distribution of topology-preserving spatiotemporal deformations, which captures realistic geometric changes in brain structures over time. These new components enable a better generation of anatomically plausible future states from an imaging scan at any time point, providing greater flexibility for modeling individual longitudinal brain trajectories.} We validate our model through both synthetic sequence generation and downstream longitudinal disease classification, as well as brain segmentation. Experiments on two large-scale longitudinal neuroimage datasets demonstrate that our method outperforms state-of-the-art baselines in generating anatomically accurate, temporally consistent, and clinically meaningful brain trajectories. Our code is available on \href{https://github.com/vfb8zv/4DLoG}{Github}.

\end{abstract}

\begin{keyword}
Longitudinal Neurodegeneration \sep Generative Diffusion Modeling \sep Deformation-Based Morphometry
\end{keyword}

\end{frontmatter}


\section{Introduction}
Neurodegenerative diseases such as Alzheimer’s disease are characterized by gradual, spatially heterogeneous atrophy of brain structures that precede clinical symptoms by years~\cite{tahami2022alzheimer,jones2022computational,thompson2007computational}. Understanding the morphometric and geometric changes of brain structures is essential for early diagnosis, individualized prognosis, and timely therapeutic intervention~\cite{gao2024identification,mofrad2021predictive}. However, longitudinal neuroimaging data are often temporally sparse, incomplete, or entirely unavailable due to challenges such as high imaging costs, patient dropout, and the difficulty of conducting repeated scans over extended study periods~\cite{mueller2005alzheimer, ortner2019amyloid}. This raises a clinically and technically significant question: can we predict an individual's future anatomical trajectory from a single baseline scan? Solving such a task would benefit early identification of high-risk individuals, simulate future disease progression, and better support clinical trials aimed at an early-stage intervention.

Recent advances in generative modeling have opened new possibilities for synthesizing future or missing brain imaging scans from limited longitudinal observations~\cite{zhu2024loci, yuan2024remind, wu2025igg}. Early generative approaches, such as generative adversarial networks (GANs)~\cite{jung2023conditional, peng2021longitudinal}, demonstrated initial promise but often suffered from instability, limited sample diversity, and mode collapse. Such issues hinder their ability to model the full variability of neurodegenerative processes. In contrast, denoising diffusion probabilistic models (DDPMs)~\cite{ho2020denoising} have recently emerged as powerful alternatives capable of capturing complex, high-dimensional distributions of brain morphology with remarkable stability and fidelity~\cite{zhu2024loci,kim2022diffusion,kim2024data}. By conditioning on auxiliary variables such as age, cognitive scores, disease status, and anatomical priors, diffusion-based frameworks have demonstrated promising results in generating high-quality, high-resolution 2D brain scans to model disease progression and structural degeneration. These advances represent a significant step toward data-driven, predictive modeling of neurodegeneration in clinical neuroscience.

\subsection{Related Works}

Current generative diffusion models synthesize brain anatomy by directly manipulating image intensities or textures, without explicitly modeling the underlying geometry of anatomical structures~\cite{zhu2024loci,kim2024data}. As a result, these methods may produce anatomically implausible samples, including unrealistic topology and structural distortions~\cite{yoon2023sadm,jayakumar2024tpie}. To address this, recent work has incorporated deformation-based morphometry (DBM) into generative frameworks~\cite{pombo2023equitable,kim2022diffusion}, representing brain changes as smooth transformations between pairwise images rather than direct intensity edits. Such transformations maintain one-to-one correspondences across time and subjects, preventing folding, tearing, or discontinuous warping~\cite{fu2025synthesizing,jayakumar2024tpie}. This property is particularly critical for studying progressive neurodegenerative diseases, where capturing subtle localized atrophy requires maintaining anatomical plausibility. Despite these advantages, existing DBM-based generative methods remain limited as they primarily model deformations between pairwise images while intermediate time points are generated via temporal interpolation~\cite{kim2022diffusion}. These approaches do not explicitly model the true longitudinal progression trajectory, thus failing to synthesize subsequent scans that reflect realistic, temporally consistent anatomical changes over several time points.

More importantly, existing generative models are fundamentally limited in their ability to process full 4D (3D × T) neuroimaging data. Most current approaches can synthesize only sequential 2D slices (2D x T)~\cite{wu2025igg}, or a single 3D follow-up volume~\cite{jayakumar2024tpie}, but they cannot generate anatomically coherent 3D sequences that evolve continuously over time. Therefore, they are unable to produce full 4D longitudinal trajectories from a single baseline scan. To alleviate this issue, several diffusion-based methods propose to synthesize follow-up scans by conditioning on a sequence of prior scans~\cite{litrico2024tadm, yoon2023sadm} or rely on auxiliary information such as image-derived features, radiomics~\cite{cho2025conditional, chintapalli2024generative}, and regional atrophy measurements~\cite{puglisi2024enhancing, ravi2019degenerative}. However, these methods assume access to multiple longitudinal scans, costly brain segmentations, or additional imaging modalities at inference time, all of which are rarely available in routine clinical workflows; hence substantially limiting their real-world applicability. Meanwhile, recent 4D diffusion models emerging in computer vision focus on synthesizing dynamic 3D scenes from fusing multiple camera viewpoints~\cite{zhang20244diffusion, liang2024diffusion4d, watsoncontrolling}. These methods are inherently tailored to multi-view consistency and camera pose estimation, making them fundamentally incompatible with the challenge of modeling biological shape changes over time in longitudinal neuroimaging.

In this paper, we propose a novel 4D longitudinal diffusion model, {\em 4DLoG}, in the time-sequential deformation space of brain images. In contrast to existing approaches~\cite{pombo2023equitable,jayakumar2023sadir} that are limited to 2D or 3D architectures - thereby sacrificing either spatial or temporal fidelity - our 4DLoG fully captures spatiotemporal anatomical dynamics by modeling a diffusion process over a sequence of 3D deformation fields. These deformation fields are parameterized by stationary velocity fields, enabling smooth, invertible transformations that preserve anatomical topology~\cite{vercauteren2008symmetric}. To support this 4D generative modeling, we introduce a new frame-wise volumetric patch embedding strategy that tokenizes each 3D volume independently while maintaining temporal consistency across the sequence. This allows us to explicitly learn the temporal evolution of brain structures without compromising spatial detail or anatomical plausibility. Our contributions are summarized below:  
\begin{itemize}
    \item \textcolor{black}{To the best of our knowledge, we are first to develop a full 4D generative diffusion model that jointly learns complete temporal sequences of 3D volumes, capturing spatial and temporal dependencies within a single model. In contrast to existing methods, our method explicitly models the longitudinal trajectory as a 4D sequence rather than generating each 3D volume independently or sequentially.}
    \item Develop a novel diffusion model in spatiotemporal deformation spaces to ensure smooth, topology-preserving transformations and anatomically consistent results.
    \item Demonstrate utility of the generated samples in two downstream tasks: AD classification and brain segmentation via augmentation with missing data from longitudinal neuroimage repositories. {\em It is worthy to note that our designed network architecture is flexible and can operate in both the intensity and deformation spaces.} 
\end{itemize}

 We evaluate the effectiveness of 4DLoG on longitudinal brain MRI data from the ADNI and OASIS repositories~\cite{petersen2010alzheimer, LaMontagne2019oasis3}. Experimental results demonstrate that our method excels at generating longitudinal sequences with preserved anatomical structure and shape, outperforming state-of-the-art diffusion models that operate in the pixel space~\cite{yoon2023sadm, puglisi2024enhancing}, as well as recursive models that synthesize missing time points using multi-frame guidance.

\section{Background: Deformation-based Brain Morphometry}
\label{sec:background}
This section outlines the formulation of topological constraints within the framework of deformation-based brain morphometry by leveraging diffeomorphic transformations between source and target brain images~\cite{miller2004computational,grenander1998computational}. In the context of longitudinal neuroimaging, it is typically assumed that any pair of scans acquired from the same subject over time can be related through a continuous deformation field~\cite{avants2008symmetric,reuter2012within,joshi2004unbiased}. To ensure anatomical fidelity and preserve the underlying brain topology, these deformation fields are constrained to lie within the space of diffeomorphisms, i.e., smooth, invertible mappings with smooth inverses~\cite{beg2005computing,arnold1966geometrie,wu2023neurepdiff}. These topological constraints are essential in DBM, as they prevent non-physical artifacts such as folding, tearing, or self-intersections in the warped anatomy, enabling reliable quantification of structural changes over time.

Given a template image $S$ and a fixed image $F$ defined on a $d$-dimensional torus domain $\Omega = \mathbb{R}^d / \mathbb{Z}^d$ ($S(x), F(x):x \in \Omega \rightarrow \mathbb{R}$), a diffeomorphic transformation, $\phi_t$, for $t \in [0, 1]$, is defined as a smooth flow over time to deform a template image to a fixed image by a composite function, $S \circ \phi^{-1}_t$. Here, the $\circ$ denotes an interpolation operator. Such a transformation is typically parameterized by time-dependent velocity fields under a large diffeomorphic deformation metric mapping (LDDMM)~\cite{beg2005computing}, or a stationary velocity field (SVF), which remains constant over time and is obtained using the scaling-and-squaring algorithm~\cite{vercauteren2008symmetric,arsigny2006log}. While we employ SVF in this paper, our framework is easily applicable to the other. For a stationary velocity field $v$, the diffeomorphisms, $\phi_t$, are generated as solutions to the equation:
\begin{equation}
\label{eq:svf}
d\phi_t / dt = v \circ \phi_t,   \,\, \text{s.t.}  \,\, \,\, \phi_0 = x.
\end{equation}
The solution of Eq.~\eqref{eq:svf} is identified as a group exponential map using a scaling and squaring scheme~\cite{arsigny2006log}. More details are included in~\cite{arsigny2006log}.

The diffeomorphic transformation, $\phi$ at $t=1$, that reflects geometric changes between images can be solved by minimizing the energy function:
\begin{equation}
\label{eq:tenergy}
E(v) = \eta ~\text{Dist}(S \circ \phi_1^{-1} (v), F) + \text{Reg}(v), \text{s.t. Eq.}~\eqref{eq:svf}. 
\end{equation}
Here Dist(·,·) is a  distance function that measures the dissimilarity between images, Reg($\cdot$) is a regularization term that enforces the smoothness of transformation fields, and $\eta$ is a positive weighting parameter. \SD{Similar to \cite{balakrishnan2018unsupervised,ashburner2007fast,beg2005computing}, this regularization term can be defined as the $L_2$-norm of the spatial gradient of the velocity field.} 
Widely used distance functions include the sum-of-squared intensity differences ($L_2$-norm)~\cite{beg2005computing}, normalized cross correlation (NCC)~\cite{avants2008symmetric}, and mutual information (MI)~\cite{wells1996multi}. In this paper, we utilize a sum-of-squared distance function. 

\section{Our Method: 4DLoG}
\begin{figure*}[!hb]
\includegraphics[width=\linewidth, trim={0.0cm 0.0cm 0.0cm 0.0cm}]{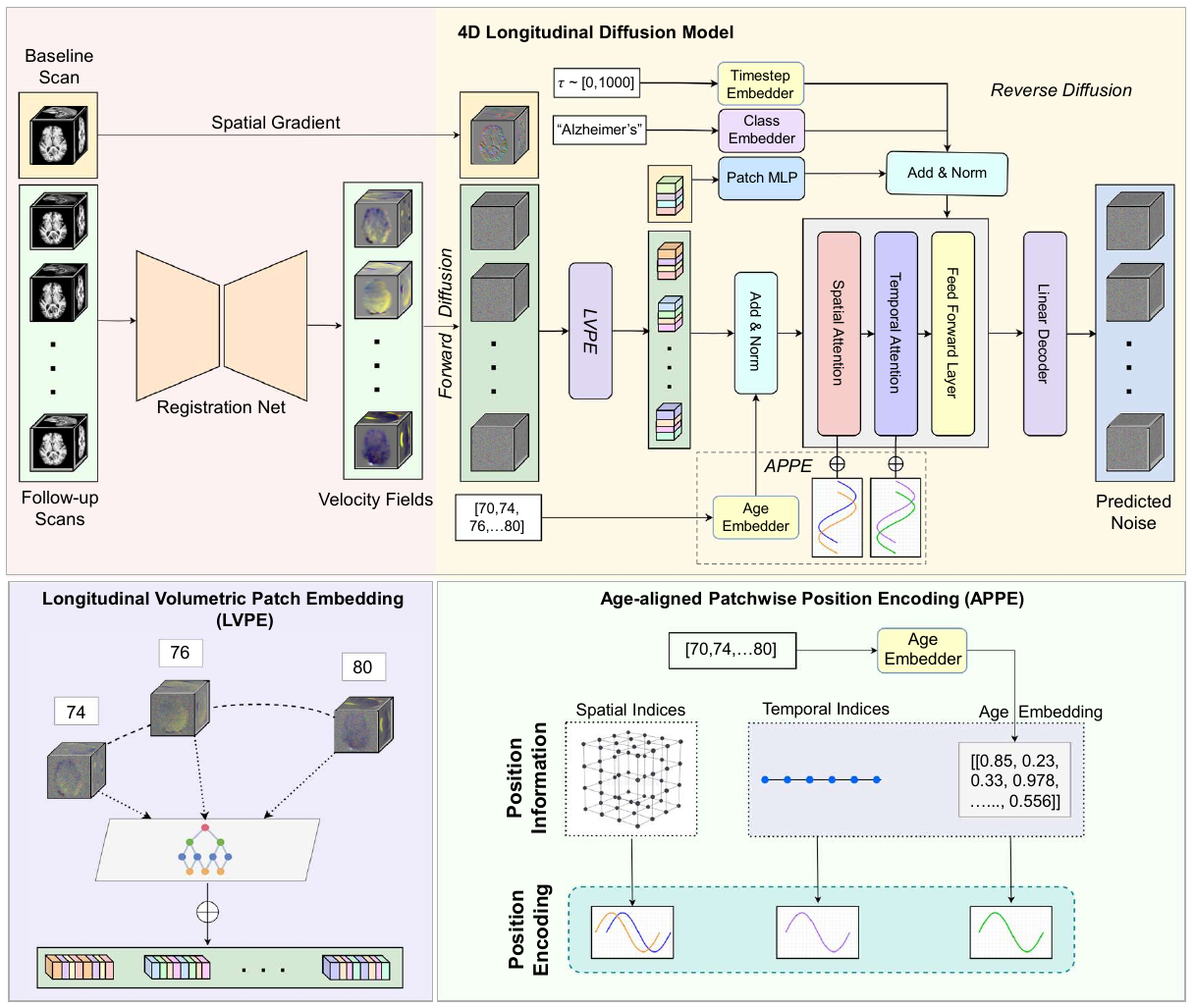}
\caption{The overall architecture of our proposed 4DLoG framework.}
\label{fig:keyfig}
\end{figure*}

This section introduces a novel longitudinal generative diffusion model, 4DLoG, which comprises of two main components: \textit{\textbf{(i)}} a diffeomorphic registration network $\mathrm{U}_\varphi$ \cite{balakrishnan2018unsupervised} that extracts DBM from the longitudinal brain images via velocity fields; and \textit{\textbf{(ii)}} a 4D diffusion model $\psi_\theta$ that learns to synthesize a time sequence of velocity fields conditioned on clinical and anatomical context. An overall network architecture is presented in Figure~\ref{fig:keyfig}.

\paragraph*{\bf Problem Setup} Different from 4D video or motion data~\cite{kim2022diffusion, liu2024texdc}, longitudinal neuroimaging data consists of discrete 3D volumetric brain scans acquired at irregular and subject-specific time points. Consider a set of $N$ longitudinal images, for each subject $n \in [1, \cdots, N]$, let $a_{t,n} \in \{ a_{0,n}, a_{1,n}, \dots, a_{T,n} \}$ denote the age at the $t$-th scan, where $T$ is the total number of time frames and $a_{0,n}$ corresponds to the baseline age. The corresponding baseline brain volume is denoted by $I_{a_{0,n}} \in \mathbb{R}^{H \times W \times L}$, where $H, W, L$ are the spatial dimensions of the 3D MRI scan. Given this baseline scan and associated clinical/demographic information such as the subject’s disease label $y_n$, our objective is to model a sequence of anatomically plausible follow-up brain volumes, $\{ I_{a_{1,n}}, I_{a_{2,n}}, \dots, I_{a_{T,n}} \}$, corresponding to subject-specific, non-uniform age intervals $\Delta a_{t,n} = a_{t+1,n} - a_{t,n}$, which may vary across individuals.

We first learn the velocity fields from the longitudinal MRIs using a diffeomorphic registration network~\cite{balakrishnan2018unsupervised}. Specifically, the baseline scan, $I_{a_0,n}$, is independently registered to each follow-up scan, $I_{a_t,n}$, resulting in a set of initial velocity fields, i.e., $\{ v_{a_{1},n}, v_{a_{2},n}, ... v_{a_T,n} \}$. These velocity fields are then integrated by Eq.~\eqref{eq:svf} to obtain smooth, invertible deformation maps $\{ \phi^{-1}_{a_{1},n}, \phi^{-1}_{a_{2},n}, ... \phi^{-1}_{a_T,n} \}$ that capture the anatomical transformations from the baseline to each follow-up time point. Let $\varphi$ denote the parameters of the registration network, we define the associated loss function as
\begin{equation*}
    l(\varphi) = \sum_{n=1}^N \lambda || I_{a_0,n} \circ \phi^{-1}_{a_t,n} \left(v_{a_t,n} (\varphi) \right)-I_{a_t,n}||_2^2 + ||\nabla v_{a_t,n} (\varphi) ||^2,
\end{equation*}
where $||\nabla v_{a_t,n}(\varphi)||^2$ is a regularizer enforcing smoothness of the transformations with $\lambda$ being a positive weighting parameter. These predicted velocities serve as training inputs for the longitudinal 4D diffusion model. \SD{Note that diffusion model captures the distribution of velocity field rather than directly modeling the deformation itself. This is a careful choice for two reasons: \textit{\textbf{(i)}} velocities are smooth and band-limited~\cite{wang2020deepflash,jia2023fourier}, which makes them a lower-dimensional and easier optimization target, \textit{\textbf{(ii)}} generating deformations directly offers no guarantee of invertibility unlike SVF~\cite{vercauteren2008symmetric,arsigny2006log} or LDDMM~\cite{beg2005computing}, hence it is prone to produce folding or tearing, whereas velocity fields map to diffeomorphisms after integration.} 

\subsection{4D Longitudinal Diffusion Model}
Inspired by denoising diffusion probabilistic models~\cite{ho2020denoising}, we develop a diffusion transformer that explicitly models longitudinal sequences within each step of the Markov chain. More specifically, our model learns to approximate the data distribution
defined over temporal sequences of velocity fields, i.e., $\mathbf{z_n} \triangleq \{v_{a_0,n}, v_{a_1,n}, ... v_{a_T,n}\}$. Specifically, each input sequence is a 4D tensor $\mathbf{z} \in \mathbb{R}^{C \times T \times H \times W \times L}$, where $C$ is the number of channels (i.e., $C=3$). For simplicity, we drop the batch notation $n$. The forward process progressively perturbs the input data over a fixed number of timesteps, transforming it into a distribution that approximates a standard Gaussian. At an intermediate timestep, $\tau \in \{1, \cdots, \Gamma\}$, the diffusion process can be formulated as  

\begin{equation} 
q(\mathbf{z}^{\tau} |  \mathbf{z}^{\tau-1}) = \mathcal{N}(\mathbf{z}^{\tau}; \sqrt{1-\beta_\tau}{\mathbf{z}^{\tau-1}}, \beta_\tau \mathbf{I}), 
\end{equation} 
where $\beta_\tau$ controls the noise variance. After re-parametrization, the last timestep of this Markov chain can be obtained as a single step process using the formulation $\mathbf{z}^\tau= \sqrt{\bar{\alpha}_\tau} \, \mathbf{z}^0 + \sqrt{1 - \bar{\alpha}_\tau} \, \epsilon$ where $\epsilon \sim \mathcal{N}(0, \mathbf{I})$, $\alpha_\tau = 1 - \beta_\tau$ and $\bar{\alpha}_\tau = \prod_{s=1}^{\tau} \alpha_s$. 

The reverse process reconstructs the signal by iteratively sampling from a Gaussian distribution using
\begin{equation}
    p_\theta(\mathbf{z}^{\tau-1} | \mathbf{z}^{\tau}) = \mathcal{N}(\mathbf{z}^{\tau-1}, \mu_\theta(\mathbf{z}^{\tau}, \tau, \mathcal{C}), \Sigma_\theta(\mathbf{z}^{\tau}, \tau, \mathcal{C})),
\end{equation}
where $\mu_\theta$ and $\Sigma_\theta$ represent the mean and variance of the process at timestep $\tau$, estimated by a network $\psi_\theta$ parameterized by $\theta$. As shown in \cite{ho2020denoising}, we can directly estimate the reverse process mean function estimator by training a neural network to predict $\epsilon$ from $\mathbf{z}_\tau$ based on a set of conditional signals $\mathcal{C}$. 

\SD{Although 4D convolutional U-Nets could in principle be applied, they scale poorly in both memory and computation, and their limited receptive fields hinder the modeling of long-range dependencies across space and time~\cite{takahashi2024comparison}. In contrast, Transformers offer a more expressive and scalable alternative, enabling global context modeling with fewer architectural constraints.} Therefore, to effectively model full 4D longitudinal sequences, we propose a new transformer-based architecture to replace the conventional U-Net denoising network, $\psi_\theta$, parameterized by $\theta$. Drawing inspiration from the video vision transformer~\cite{arnab2021vivit}, we redesign the denoising backbone with dedicated components optimized for spatiotemporal modeling, as described below: 

\paragraph{\bf Longitudinal Volumetric Patch Embedding (LVPE)} Existing methods generate longitudinal data by processing sequences of 3D volumes \cite{yoon2023sadm, liu2024texdc}; however, they often compromise either spatial or temporal coherence by flattening or reshaping the input sequence to make it compatible to current transformer architectures. To address this, we introduce a patch extraction mechanism that operates in the 4D spatiotemporal domain. \SD{The reason for such a design is two fold: \textit{\textbf{(i)}}  brain anatomy within a single scan is spatially correlated and features extracted via convolution preserves intrinsic features better as compared to unfolding into patches, \textit{\textbf{(ii)}} frame-wise embedding avoids the assumption of uniform temporal spacing required by spatiotemporal or tubelet-based embeddings~\cite{zhao2022tuber,singh2022video}, making our proposed design suitable for irregular longitudinal data.} As a result, we first apply a 3D convolutional-based patch embedder $\kappa$ independently to each temporal frame, as brain anatomical topology remains fundamentally stable across time points with minor deformations occurring between scan intervals. We then partition each 3D volume into non-overlapping patches and project them into a fixed-dimensional embedding space $\mathbb{R}^d$ of dimension $d$, yielding a patch token tensor of shape $m \in \mathbb{R}^{T \times N_d \times d}$, where $N_d=(H \times W \times L)/P^3$ is the number of spatial patches per frame with a patch size of $P$. Formally, the embedding process is defined as
\begin{align*}
m_{a_t} &= \kappa(\mathbf{z}_{a_t}), \quad \forall a_t \in \{a_1, \dots, a_T\}, \\
m &= [m_{a_1} \, m_{a_2} \, \cdots, \, m_{a_T}] \, ,\, m \in \mathbb{R}^{T \times N_d \times d}.
\end{align*}
Note that our proposed operation differs from  current works~\cite{peebles2023scalable,ma2024latte} by incorporating 4D data through frame-wise 3D patch embeddings. This approach preserves the fine-grained spatial structure within each 3D volume while maintaining temporal correspondence across time points, enabling effective decoupling of spatial and temporal modeling in subsequent transformer layers. In contrast to tubelet-based embeddings~\cite{zhao2022tuber,singh2022video} that extract spatiotemporal tubes assuming uniform temporal spacing, our frame-wise patch extraction is naturally suited to longitudinal medical data with irregular age intervals $\Delta a_{t}$. This design choice avoids the need for temporal interpolation or padding, reducing computational overhead while improving modeling flexibility. 

\paragraph{\bf Age-aligned Patchwise Position Encoding (APPE)} The patches extracted from the input sequence are enriched with both spatial and temporal positional encodings to preserve temporal alignment and to capture global spatiotemporal dependencies. A key innovation of our design is a two-step temporal alignment strategy. {\em First}, we introduce a temporally aware encoding that embeds age information directly into the sinusoidal positional functions, enabling the model to reason about continuous biological time. {\em Second}, we incorporate a complementary fixed 1D temporal embedding assigned to each volumetric patch, providing a stable temporal reference across the sequence. To ensure that the ages are accurately incorporated with their respective 3D volumes, we define linear temporal position encoding using sinusoidal embeddings, followed by a non-linear multilayer perceptron (MLP) transformation. Each linear age embedding $f(a_t)$ is defined as a continuous function, i.e.,
\begin{align*}
    f(a_t) = [(cos(\omega_{i'}.a_t))_{{i'}=0}^{\frac{\mathbf{D}}{2}-1}, (sin(\omega_{i'}.a_t))_{{i'}=0}^{\frac{\mathbf{D}}{2}-1}] \in \mathbb{R}^d,
\end{align*}
where $\omega_{i'} = 1/{10000^{\frac{2i'}{\mathbf{D}}}}$. The age-aligned temporal encoding, MLP$(f(a_t))$, is then added to the extracted patches along the temporal axis. In addition, we introduce a second set of fixed temporal sinusoidal encodings, which are injected before each temporal attention block in the transformer. This dual-encoding strategy provides both biologically grounded temporal context and a stable temporal reference to improve the model’s ability to learn coherent spatiotemporal dynamics.

We then employ a fixed 3D sinusoidal spatial positional encoding that assigns each token a unique embedding based on its location in a 4D spatiotemporal grid. Inspired by~\cite{ma2024latte}, we construct independent 1D sinusoidal encodings along each spatial axis and combine them into a full 3D positional embedding. This design guarantees a unique representation for every spatial coordinate in the volume while enabling the transformer to capture long-range, global spatial dependencies. Beyond being transformer-agnostic, this approach also overcomes the limitations of relative or axis-decoupled encodings, which cannot fully encode multidimensional spatial context and may inadvertently assign identical embeddings to distinct patches with the same relative index.

\paragraph{\bf Adaptive Spatio-Temporal Contextualization using Anatomical Embeddings} Our design is motivated by the need to incorporate both subject-specific anatomical context and temporally coherent disease progression, which standard transformer architectures and existing conditioning schemes fail to fully capture. We propose a multimodal conditioning mechanism that integrates both disease class and anatomical priors directly into the transformer. Specifically, the anatomical prior is defined as the spatial gradient of the initial image scan, $\nabla I_{a_0}$, which aligns with the directionality of learned transformations~\cite{beg2005computing, joshi2000landmark}. \SD{The disease label 
$y$ and diffusion timestep $\tau$ are embedded through separate MLP blocks as Class Embedder and Timestep Embedder respectively. These consist of learnable non-linear MLP layers. The anatomical prior is embedded through the patch extractor followed by its separate MLP layers (see Fig \ref{fig:keyfig}). The resulting embeddings are averaged to form a global conditioning vector and injected into the normalization layers of each DiT block via adaptive layer normalization \cite{peebles2023scalable,guo2022adaln}. This layer predicts the scale and shift parameters to replace the parameters of standard Layer-Normalization layer.} This avoids the overhead and complexity of voxel-wise cross-attention mechanisms, which often require explicit spatial alignment—an unrealistic assumption in longitudinal synthesis where anatomical structure evolves across time. To jointly capture anatomical detail and temporal consistency, we employ a factorized space-time attention scheme~\cite{arnab2021vivit} to alternate between spatial and temporal transformer blocks. Spatial blocks attend to 3D patches within each timepoint, preserving intra-frame anatomical structure, while temporal blocks attend across frames at fixed spatial locations, modeling age-dependent progression. This enables anatomy-aware attention for individual frames and temporally coherent synthesis across the sequence, which are crucial for realistic 4D generation.

\subsection{Training Objective} 
Given a sequence of clean initial latent features $(\mathbf{z}^0)$ and a randomly sampled timestep \(\tau \in \{1, \ldots, \Gamma\}\), we train the model to predict the added noise, $\epsilon$, from $\mathbf{z}^{\tau}_{a_t} = \sqrt{\bar{\alpha}_\tau} \mathbf{z}^{0}_{a_t} + \sqrt{1 - \bar{\alpha}_\tau} \epsilon$. The denoiser $\psi_\theta(\mathbf{z}^{\tau}, \tau, \mathcal{C})$ is conditioned on diffusion timestep $\tau$ and additional condition $\mathcal{C}$ (i.e., age, baseline scan and disease), and is trained to minimize the $\mathrm{L1}$ error
\begin{align}
\mathcal{L}_{\text{diff}} = \mathbb{E}_{z^{\tau}, \epsilon, \tau} \left[ \left\| \epsilon - \epsilon_\theta(\mathbf{z}^{\tau}, \tau, \mathcal{C}) \right\|_1 \right].
\end{align}
This formulation encourages our model to learn a conditional score function that reverses the forward noise process, generating consistent longitudinal velocity fields. These velocity fields are then integrated to deform the baseline scan, producing a temporally ordered 4D trajectory of brain anatomy. 

The training and sampling of 4DLoG are outlined in Alg.~\ref{alg:training} and Alg.~\ref{alg:sampling}.
\begin{algorithm}[!htbp]
\caption{4DLoG Training}
\label{alg:training}
\textbf{Inputs:} Scans $\{I_{a_0}, I_{a_1}, \ldots, I_{a_T}\}$, ages $[a_0, a_1,...,a_T]$, disease class label $y$ 
\textbf{Stage 1: Registration Net $U_\varphi$ Training}
\begin{algorithmic}[1]
\For{$n \in [1,N]$}
    \State Template $\leftarrow$ $I_{a_0,n}$ ; Fixed $\leftarrow$ $\{I_{a_1,n}, \ldots, I_{a_T,n}\}$.
    \For{$a_{t,n} \in [a_{1,n}, a_{T,n}]$}
    \State $v_{a_t,n} \leftarrow {U}_\varphi(I_{a_0,n}, I_{a_t,n})$ 
    \EndFor
    \State Minimize $l(\varphi)$
\EndFor
\end{algorithmic}

\textbf{Stage 2: 4D Longitudinal Diffusion Model $\psi_\theta$ Training}
\begin{algorithmic}[1]
\State $\mathbf{z_n^0} \leftarrow [v_{a_1,n},...,v_{a_T,n}] $ (dropping $'n'$)
\State $\mathbf{z^\tau} \leftarrow \sqrt{\bar{\alpha}_\tau} \, \mathbf{z}^0 + \sqrt{1 - \bar{\alpha}_\tau} \, \epsilon$
\State $m = [\kappa(\mathbf{z_{a_1}^\tau}), \kappa(\mathbf{z_{a_2}^\tau}),...,\kappa(\mathbf{z_{a_T}^\tau})]$
\State $m \leftarrow [(m_{a_1}+f(a_1)),...,(m_{a_T}+f(a_T))]$
\State $m \leftarrow m +$ spatial encoding. 
\For{$\hat{\mathrm{l}} \in \# \text{layers}$} 
\State $m \leftarrow \textit{Spatial Block(m)}$ 
\State $m \leftarrow m+$ temporal encoding. 
\State $m \leftarrow \textit{Temporal Block(m)}$ 
\State $m \leftarrow \textit{Normalization}(m,\kappa(\nabla I_{a_0}),\tau_{embed},y_{embed})$
\EndFor
\State $\epsilon_\theta \leftarrow Einsum(m)$
\State Minimize $\mathcal{L}_\mathrm{diff}$
\end{algorithmic}

\end{algorithm}

\begin{algorithm}[!htbp]
\caption{4DLoG Sampling}
\label{alg:sampling}
\textbf{Inputs:} Predictor step, $\Gamma = 1000$, Corrector step, $M=2$
\begin{algorithmic}[1]
\State \textbf{Initialize} $\mathbf{z}_\Gamma \sim P_{\Gamma}(x)$ 
\For{$\tau= \Gamma-1,\ldots,0$}
    \State $\mathbf{z}_\tau \leftarrow \text{Predictor}(\mathbf{z}_{\tau+1})$
    \For{$\hat{\tau} = 1,\ldots,M$}
        \State $\mathbf{z}_\tau \leftarrow \text{Corrector}(\mathbf{z}_\tau)$
    \EndFor
\EndFor
\State \Return $\mathbf{z}_0$
\end{algorithmic}
\end{algorithm}

\section{Experiments}
\textcolor{black}{This section details the validation of our proposed 4DLoG model on publicly available longitudinal brain MRI datasets, Alzheimer's Disease Neuroimaging Initiative (ADNI) and Open Access Series of Imaging Studies (OASIS)~\cite{petersen2010alzheimer, LaMontagne2019oasis3}. We first compare its performance with state-of-the-art baselines~\cite{puglisi2024enhancing, pombo2023equitable, yoon2023sadm}, evaluate generation quality using quantitative metrics and assess the statistical significance of differences in performance using paired t-tests. We then provide qualitative comparisons to demonstrate the ability of 4DLoG to capture longitudinal changes. The model's performance is further evaluated on two downstream tasks, disease classification on ADNI and hippocampus segmentation on OASIS, using data augmentation with generated missing longitudinal scans for each patient. These tasks provide an additional measure of whether the proposed model generates clinically meaningful longitudinal scans.}

\subsection{Dataset.} 
\SD{We utilize longitudinal brain MRI data from two public repositories~\cite{petersen2010alzheimer,LaMontagne2019oasis3} to train our framework and validate its performance via downstream classification and segmentation tasks. Both datasets contain clinical disease labels for cognitively normal (CN), mild cognitive impairment (MCI), and Alzheimer’s disease (AD) subjects. Our 4DLoG model is trained and evaluated on ADNI, conditioned on age and disease diagnosis. For the downstream tasks, 4DLoG is used to generate missing longitudinal scans for subjects from both ADNI and OASIS cohorts. The neuroanatomical segmentation labels for OASIS~\cite{LaMontagne2019oasis3} are extracted via SynthSeg~\cite{billot2023synthseg}.  Details of dataset composition, temporal characteristics, and preprocessing steps are provided below.}

\paragraph*{\bf ADNI \cite{petersen2010alzheimer}} We use T1-weighted  MRIs of $1021$ participants with at least 4 longitudinal visits from the ADNI repository \cite{mueller2005alzheimer}. All subjects had corresponding disease diagnoses and age information at their observed time points. All scans are skull-stripped, intensity normalized, and affine-registered to a common template space~\cite{brett2002problem}. Based on the disease diagnosis at the time of visit, the subjects (aged $55 - 92$) are divided into three classes - Cognitively Normal (CN), Alzheimer's Disease (AD), and Mild Cognitive Impairment (MCI). Each sample is resized to ($128\times128\times128\times4$), i.e., 3D volumes at 4 time points. Metadata includes age and diagnosis at each individual visit. The dataset is split into $70\%/15\%/15\%$ training, validation and testing, with each subject having at least $4$ time-points. 

\textcolor{black}{For the downstream classification task, we randomly sampled $512$ subjects from ADNI, covering the three aforementioned disease classes (AD, MCI, CN) with $181/201/ 130$ subjects per class. }

\paragraph*{\bf OASIS~\cite{LaMontagne2019oasis3}} \textcolor{black}{The OASIS3 dataset is used solely to evaluate the downstream segmentation task. We select $53$ subjects with at least $4$ longitudinal T1-weighted scans, yielding $212$ 3D volumes. Data is split at the subject level into $42$ for training and $11$ for testing. All scans are skull-stripped, bias-field corrected, and intensity normalized to the $[0,1]$ range. To ensure compatibility with our registration framework trained on ADNI, we affine-registered each volume to the ADNI template space using the ANTs toolkit~\cite{tustison_antsx_2021} and resampled to a common resolution of $128^3$, matching the ADNI preprocessing.} \textcolor{black}{Following affine alignment, we obtain segmentation labels for the hippocampus for each volume using SynthSeg~\cite{billot2023synthseg} within the FreeSurfer framework. We focus on hippocampal segmentation since hippocampal atrophy is a well-established imaging biomarker strongly associated with Alzheimer’s disease~\cite{barnes2009meta} and progresses more prominently.  These segmentations are subsequently used as the reference standards for training and evaluating the baseline segmentation model.} 

\subsection{Experimental Setup}
We evaluate 4DLoG by assessing the quality, fidelity, and anatomical consistency of the synthesized longitudinal MRI scans. This includes comparison against state-of-the-art baselines, analyses across multiple quantitative metrics, and ablation studies to study the effect of model capacity, temporal conditioning, and autoencoders for various synthesis strategies. We further assess the utility of the generated scans for downstream tasks, including classification and segmentation, to demonstrate their clinical applicability and value in longitudinal MRI analysis. \\

\noindent\textbf{Baselines.} We compare our method (of both intensity- and deformation-based variants) against three state-of-the-art diffusion-based baselines for longitudinal MRI generation: Sequence-Aware Diffusion Model (SADM) \cite{yoon2023sadm}, BrLP~\cite{puglisi2024enhancing} and CounterSynth~\cite{pombo2023equitable}. While several prior works focus on interpolating between two scans~\cite{kim2022diffusion}, our evaluation focuses on comparison with models that extrapolate future volumes to promote a fair benchmark given that our method relies solely on the baseline scan.

\textcolor{black}{Please note that SADM~\cite{yoon2023sadm} requires multiple follow-up scans as input to synthesize a single volume, making it less practical. Due to its high computational demands, we implement a latent version using features from a pretrained autoencoder \SD{similar to \cite{puglisi2024enhancing}}. \SD{Specifically, we utilize the same 3D variational autoencoder (VAE) as our framework in the intensity space, which we train with a generative adversarial loss and perceptual loss, in addition to a vanilla VAE’s reconstruction and KL divergence objectives \cite{pinaya2022brain, rombach2022high}. The latent dimension is set as $3\times32^3$. The patch size is set to $8^3$ to accommodate the latent dimension, while the rest of the parameters are left unchanged from the original implementation.}}

\textcolor{black}{Both BrLP~\cite{puglisi2024enhancing} and CounterSynth~\cite{pombo2023equitable} are 3D generative models that synthesize individual scans independently in image spaces, rather than explicitly modeling longitudinal sequences or learning temporal dependencies across time points. BrLP can recurrently generate follow-up volumes from a single baseline scan; however, its robust performance relies on additional regional segmentation masks as auxiliary inputs. To ensure a fair comparison under the same experimental setting, we use the BrLP components trained with FreeSurfer segmentation labels but generate subsequent volumes without the auxiliary (AUX) component, since ground-truth segmentation labels are not provided to our model.} 

\noindent\textbf{Evaluation Metrics.} Conditioned on baseline scan, age, and diagnosis, we evaluate the generated followup MRI scans using multiple metrics to assess both accuracy and realism. For accuracy, we employ Peak Signal-to-Noise Ratio (PSNR)~\cite{eskicioglu2002image} and Structural Similarity Index Measure (SSIM)~\cite{wang2004image}, both standard in image quality assessment. To measure realism, we compute the Fréchet Inception Distance (FID) and Kernel Inception Distance (KID) in the feature space, following established protocols and using a standard pre-trained model~\cite{chen2019med3d}. We also evaluate the anatomical consistency of our generated deformation fields by analyzing the Determinant of Jacobian (DetJac) across frames and LDT model variants. Low percentages of negative DetJac values indicate preserved topology, providing a measure of how reliably the synthesized scans maintain anatomically plausible deformations. \\

\noindent\textbf{Downstream Task.} We evaluate the utility of our synthesized longitudinal scans as a data augmentation strategy for downstream classification and segmentation. This allows us to assess whether the synthetic images can effectively supplement real data, particularly in settings where labeled samples are limited. \SD{The downstream validation considers subjects with missing intermediate scans and available ground-truth disease labels at all observed time points. We assume that the disease status remains consistent throughout the longitudinal sequence, with the final-stage diagnosis matching the baseline diagnosis. We synthesize the missing scans and incorporate them into the original training set for data augmentation, followed by downstream evaluation to determine whether the generated scans capture disease-relevant longitudinal information.}

\SD{For classification, we apply the trained 4DLoG to synthesize missing longitudinal scans for subjects with fewer than four available time points, including $176$ subjects with one available scan, $188$ subjects with two available time points, and $148$ subjects with three available time points. These synthesized scans covers $130/201/181$ healthy/MCI/AD, and are used to augment the original dataset of $512$ baseline scans, with a train/validation/test split of $70\%/15\%/15\%$. Because the missing time points also lack age information, we sample the ages of the synthesized scans from group-specific normal distributions with the means and standard deviations estimated from the training set. We then incorporate these synthesized scans into the original training set and use it to train the downstream VGG3D classifier~\cite{rahman2025alzheimer}. This allows us to assess whether longitudinal data synthesized by 4DLoG can improve disease classification performance.}

\textcolor{black}{For segmentation, we first train a baseline 3D U-Net model~\cite{ronneberger2015u} using the initial scans and the corresponding segmentation labels from the original OASIS longitudinal dataset. We then apply 4DLoG to synthesize the subsequent missing scans for each subject, progressively incorporating the generated scans to augment the training set. Importantly, we apply the same synthesized deformations generated by 4DLoG to both the images and their corresponding segmentation maps to ensure that the synthesized image–segmentation pairs remain anatomically consistent across the longitudinal sequence. For each augmentation setting, the resulting dataset is randomly split into 80\% for training and 20\% for testing.} \\

\noindent\textbf{Ablation Studies.} Our ablation studies focus on model architecture, temporal conditioning, and synthesis strategies. We first experiment with three transformer variants (LDT-S, LDT-L, and LDT-XL) with different number of hidden embedding dimensions, transformer blocks, and attention heads, to study the scalability of our deformation-based model (4DLoG-Def.). 

We then compare two age-conditioning strategies: age-wise temporal position encoding (APPE) applied to extracted volumetric patches, and age vectors incorporated through adaptive normalization alongside diffusion timestep, disease class, and anatomical prior. 

Finally, we implement an intensity-based version (4DLoG-Int.) by replacing the registration network with a latent-space VAE-GAN~\cite{pinaya2022brain, rombach2022high} and compare it against our deformation-based model (4DLoG-Def.) that learns the distribution of velocity fields from a pre-trained registration network. These studies allow us to analyze the effect of model capacity, temporal encoding, and synthesis approach on generating anatomically plausible longitudinal scans.

\subsection{Implementation Details.} 
Similar to DDPM \cite{ho2020denoising}, we set the total number of diffusion time steps as $1000$. A cosine noise schedule~\cite{nichol2021improved} is used in the diffusion process. \textcolor{black}{We fix the number of time points to $4$ to accommodate the substantial variability in the number of longitudinal scans available across subjects. The diffusion process is applied to the complete 4D sequence, with diffusion parameters, including the per-time-point mean and added noise, defined over the full temporal dimension rather than being independently or repeatedly applied to each volume. Disease labels are available for all observed time points, and we assume that the disease status remains consistent throughout each longitudinal sequence due to the limited samples of transited subjects. We use a 70/15/15\% train/validation/test split with subject-level separation to ensure that no subject appears across different splits, and select the final model based on its performance on the validation set.}

\textcolor{black}{For the LVPE module, we use a patch size and stride of $4$, with zero padding.} All networks are trained with a learning rate of $10^{-4}$, effective batch size of $48$ and the Adam optimizer. We train the registration network for $1500$ epochs and the diffusion model for approximately $200$K training steps. While the final brain MRIs are at a resolution of $128^3$, our diffusion model synthesizes velocity fields at a lower resolution of $32^3$ for computational efficiency. Since velocities exhibit a smooth, band-limited structure~\cite{wang2020deepflash,jia2023fourier}, we up-sample the synthesized velocity field back to the original resolution using trilinear interpolation. All experiments are conducted on NVIDIA A100 GPUs.

\section{Results}
\subsection{Evaluation of Sample Fidelity. }
\begin{figure*}[!h]
\centering
 \includegraphics[width=\textwidth,height=\textheight,keepaspectratio]{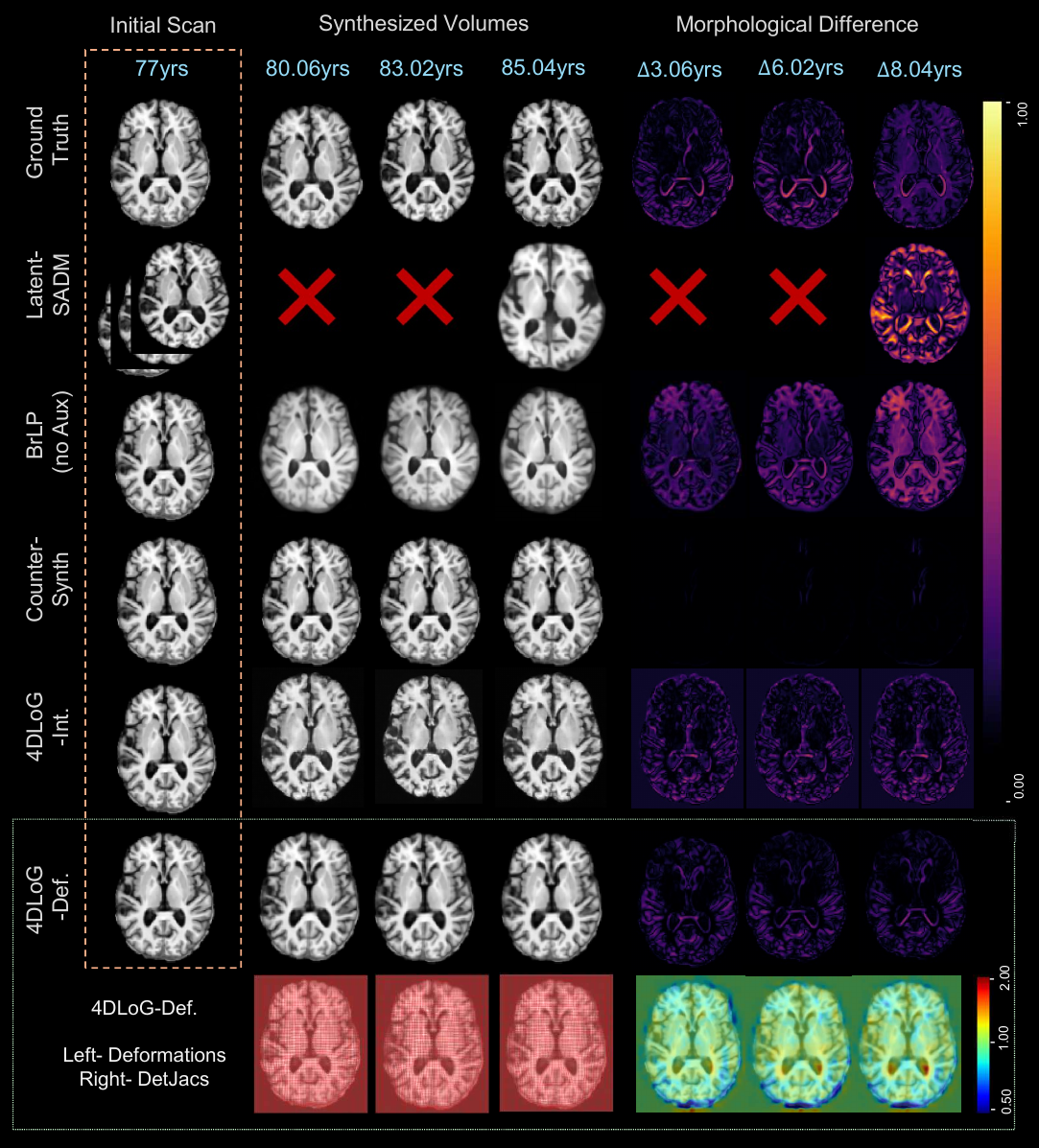}
\caption{\textcolor{black}{Left to right: Comparison of synthesized follow-up volumes across all methods, along with morphological difference maps that highlight longitudinal changes from the initial scan for a subject with {\bf Cognitively Normal} (shown in the {\bf axial view}). For our proposed deformation-based model, we additionally visualize the estimated deformation field and the corresponding Jacobian determinant (DetJac) that reflect topological structure of the brain changes over time.}}
\label{fig:cn_sota_ax}
\end{figure*}

\begin{figure*}[!h]
\centering
 \includegraphics[width=\textwidth,height=\textheight,keepaspectratio]{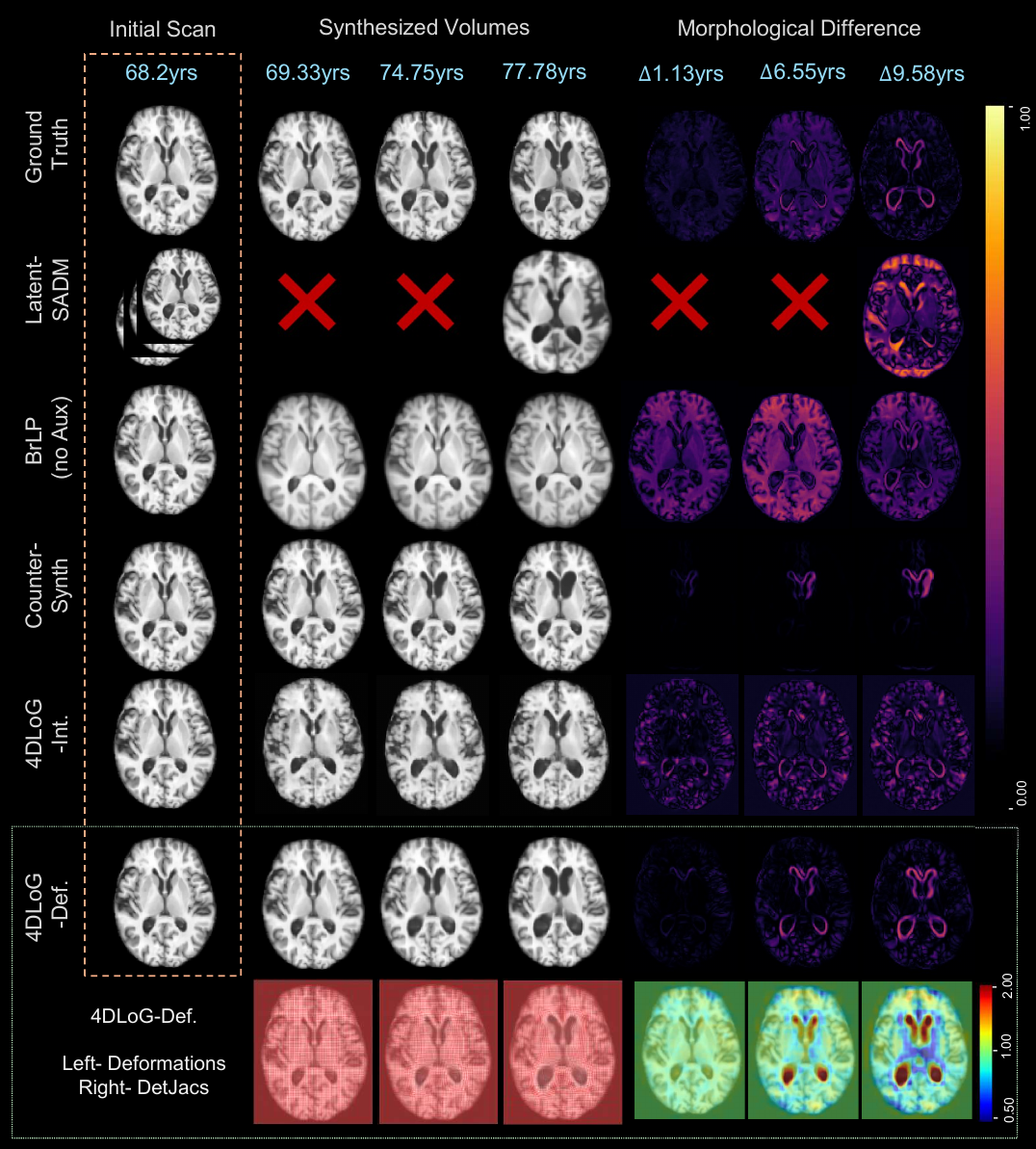}
\caption{\textcolor{black}{Left to right: Comparison of synthesized follow-up volumes across all methods, along with morphological difference maps that highlight longitudinal changes from the initial scan for a subject with {\bf Mild Cognitive Impairment} (shown in the {\bf axial view}). For our proposed deformation-based model, we additionally visualize the estimated deformation field and the corresponding Jacobian determinant (DetJac) that reflect topological structure of the brain changes over time.}}
\label{fig:mci_sota_ax}
\end{figure*}

\begin{figure*}[!h]
\centering
 \includegraphics[width=\textwidth,height=\textheight,keepaspectratio]{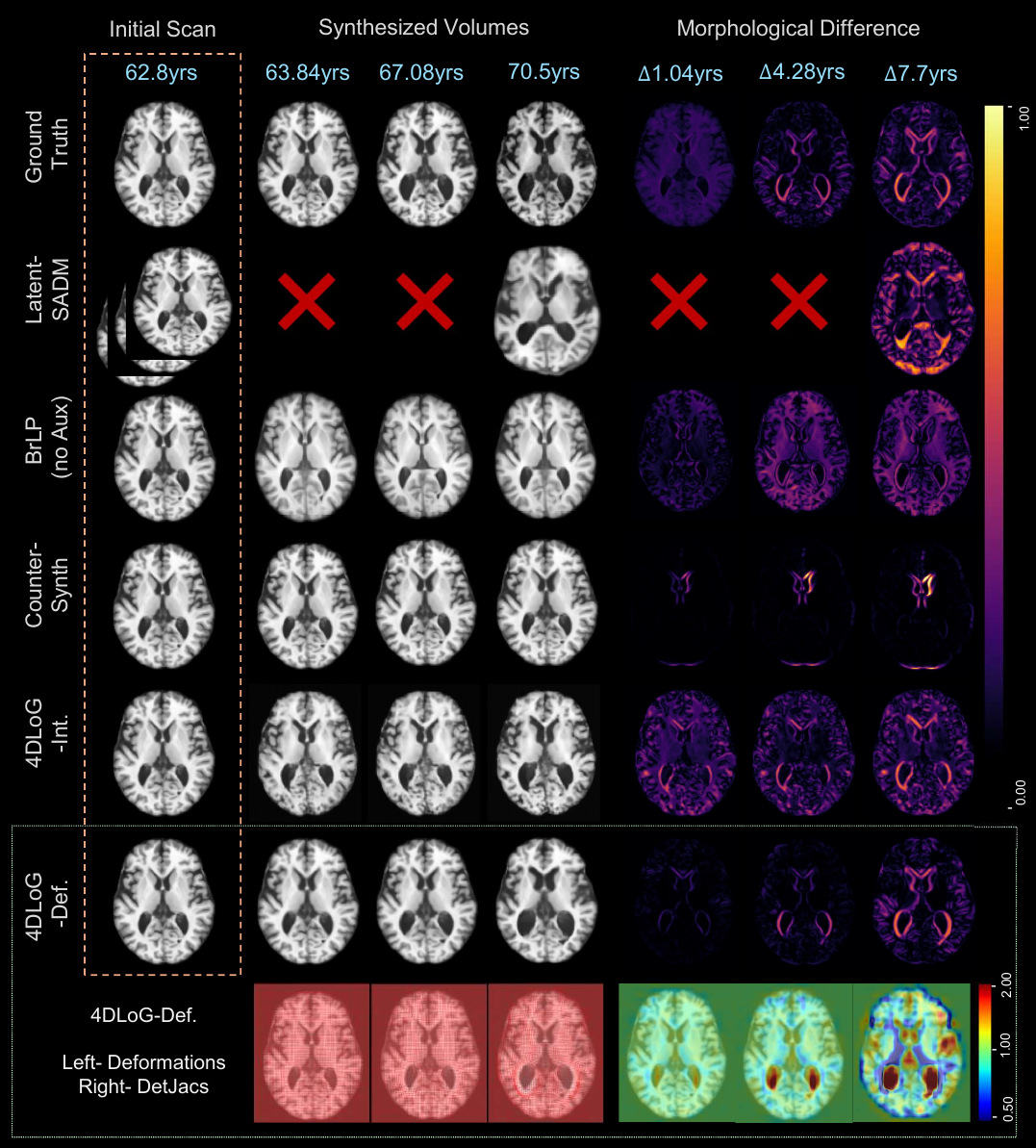}
\caption{\textcolor{black}{Left to right: Comparison of synthesized follow-up volumes across all methods, along with morphological difference maps that highlight longitudinal changes from the initial scan for a subject with {\bf Alzheimer’s Disease} (shown in the {\bf axial view}). For our proposed deformation-based model, we additionally visualize the estimated deformation field and the corresponding Jacobian determinant (DetJac) that reflect topological structure of the brain changes over time.}}
\label{fig:main_results}
\end{figure*}

Figures \ref{fig:cn_sota_ax}, \ref{fig:mci_sota_ax}, and \ref{fig:main_results} illustrate examples of anatomical changes of CN, MCI, and AD captured by our framework compared to SOTA methods. Visually, Latent-SADM~\cite{yoon2023sadm} fails to preserve anatomical structures as the age gap increases, while BrLP~\cite{puglisi2024enhancing} and CounterSynth~\cite{pombo2023equitable} fail to model the progression of cortical degeneration seen in subjects with AD. In contrast, both our models (4DLoG-Int./Def.) generate realistic changes in brain volume, conditioned on ages. Please note that while our intensity-based variant, 4DLoG-Int., occasionally introduces anatomically implausible regions, the deformation model, 4DLoG-Def., consistently preserves brain topology and accurately models longitudinal progression. These results also highlight that our model better preserves structural integrity and more effectively captures the anatomical progression patterns across the cognitive spectrum of CN, MCI, and AD. Notably, it reflects the accelerated atrophy characteristic of AD, including prominent ventricular enlargement and visible shrinkage of both gray and white matter structures. In contrast, MCI shows more gradual structural changes, while CN remains largely stable over time. 

To further quantitatively verify anatomical consistency, we also compute the determinant of Jacobian (DetJac) distributions for our deformation-based model \textcolor{black}{(as shown on the bottom panel of Figures~\ref{fig:cn_sota_ax}, ~\ref{fig:mci_sota_ax}, ~\ref{fig:main_results})}. DetJac serves as a standard measure of topological preservation, where the values of $1$ indicate local volume preservation, values $<1$ denote shrinkage, and values $>1$ indicate expansion. Crucially, negative DetJac values correspond to anatomically implausible transformations, such as folding or singularities, that violate topology. DetJac of our deformation model accurately captures tissue atrophy and cerebrospinal fluid expansion with merely $3.5\times10^{-4}\%$ negative values across all samples, demonstrating its strong ability to maintain anatomical consistency.

Table~\ref{tab:my_label} presents the quantitative comparison of all methods, \textcolor{black}{with the mean and standard deviation for each metric across all time points. It shows that our deformation-based model (4DLoG-Def.) achieves the best performance on all four metrics, which indicates better fidelity to the real data distribution. Additionally, the higher PSNR and SSIM further show that the synthesized volumes are accurate at the voxel and structural level.}

\textcolor{black}{Table ~\ref{tab:pvalues} shows the p-values from paired t-tests between our 4DLoG Def. method against each baseline. Overall, it shows that our proposed models (intensity or deformation-based) achieve superior fidelity across all metrics. Even against the closest baseline, CounterSynth, paired $t$-tests show that the improvements achieved by our method are statistically significant ($p<0.05$) for pairwise PSNR and SSIM, as well as disease-specific FID and KID across all three time points.}

\begin{table*}[!hb]
    \centering
    \setlength{\tabcolsep}{4pt} 
    \begin{tabular}{lccccc}
        \toprule
         Metric & \color{black} SADM & \color{black} BrLP & CounterSynth & 4DLoG-Int. & 4DLoG-Def.  \\
        \midrule
        Input & \color{black} $\{I_{a_i}\}_{i=0}^{T-1}$ & \color{black} $I_{a_{0}}$ & $I_{a_{t-1}}$ & $I_{a_{0}}$ & $I_{a_{0}}$ \\
        Output & \color{black} $I_{a_T}$ & \color{black} $I_{a_{t}}$ & $I_{a_t}$ & $\{I_{a_i}\}_{i=1}^{T}$ & $\{I_{a_i}\}_{i=1}^{T}$  \\
        \midrule
        FID $\downarrow$ & \color{black} $9.255$ \scriptsize{$\pm 0.00$} & \color{black} $2.095$ \scriptsize{$\pm 0.028$} & $0.78$ \scriptsize{$\pm 0.57$} & $1.014$ \scriptsize{$\pm 0.812$} & $\mathbf{0.241}$ \scriptsize{$\pm 0.2$} \\
        PSNR $\uparrow$ & \color{black} $19.16$ \scriptsize{$\pm 1.71$} & \color{black} $22.38$ \scriptsize{$\pm 2.81$} & $25.39$ \scriptsize{$\pm 2.05$} & $24.72$ \scriptsize{$\pm 1.73$} & $\mathbf{25.90}$ \scriptsize{$\pm 1.30$}\\
        SSIM $\uparrow$ & \color{black} $0.83$ \scriptsize{$\pm 0.012$} & \color{black} $0.87$ \scriptsize{$\pm 0.044$} & $0.92$ \scriptsize{$\pm 0.02$} & $0.89$ \scriptsize{$\pm 0.01$} & $\mathbf{0.93}$ \scriptsize{$\pm 0.01$} \\
        KID $\downarrow$ & \color{black} $1.119$ \scriptsize{$\pm 0.00$} & \color{black} $0.053$ \scriptsize{$\pm 0.001$} & $0.008$ \scriptsize{$\pm 0.008$} & $0.037$ \scriptsize{$\pm 0.04$} & $\mathbf{0.003}$ \scriptsize{$\pm 0.005$}\\
        \bottomrule
    \end{tabular}
    \color{black} \caption{Quality of generated MRI scans by different methods based on ADNI.}
    \label{tab:my_label}
\end{table*}

\begin{table*}[!ht]
    \centering
    \color{black}
    \setlength{\tabcolsep}{16pt}
    \begin{tabular}{lcccc}
        \toprule
        Model & FID & PSNR & SSIM & KID \\
        \midrule
        SADM & $0.035$ & $3.64e\text{-}42$ & $2.07e\text{-}80$ & $0.037$ \\
        BrLP        & $2.40e\text{-}5$ & $4.22e\text{-}27$ & $2.24e\text{-}37$ & $3.10e\text{-}3$ \\
        CounterSynth & $1.13e\text{-}3$ & $1.18e\text{-}17$ & $2.56e\text{-}9$ & $1.34e\text{-}3$ \\
        \bottomrule
    \end{tabular}
    \caption{P-values from paired t-tests between each baseline and 4DLoG-Def.}
    \label{tab:pvalues}
\end{table*}

Table~\ref{tab:detjac_summary} reports the mean, standard deviation, and percentage of negative values in the Determinant of Jacobian (DetJac) of the deformation fields across frames and LDT model variants. A low or zero percentage of negative values indicates preservation of topology in the generated deformation fields. As seen in our results, all variants of our model exhibit a notably low percentage of negative DetJac values, indicating consistent preservation of topology in the synthesized scans.
\begin{table}[!htbp]
    \centering
    \footnotesize
    \setlength{\tabcolsep}{12pt}
    \begin{tabular}{c l c c c}
        \toprule
        Model & Frame & Mean & \makecell{Std.Dev.} & $-\text{DetJac}\% \downarrow$ \\
        \midrule
        \multirow{3}{*}{\rotatebox[origin=c]{0}{LDT-S}} 
            & Frame 1 & $1.000$ & $0.043$ & $1.892e\text{-}6$ \\
            & Frame 2 & $0.999$ & $0.047$ & $0.0000$ \\
            & Frame 3 & $0.999$ & $0.058$ & $0.0000$ \\
        \midrule
        \multirow{3}{*}{\rotatebox[origin=c]{0}{LDT-L}} 
            & Frame 1 & $1.000$ & $0.045$ & $1.298e\text{-}4$ \\
            & Frame 2 & $1.000$ & $0.050$ & $7.466e\text{-}4$ \\
            & Frame 3 & $0.999$ & $0.055$ & $1.869e\text{-}4$ \\
        \midrule
        \multirow{3}{*}{\rotatebox[origin=c]{0}{LDT-XL}} 
            & Frame 1 & $0.999$ & $0.102$ & $8.250e\text{-}5$ \\
            & Frame 2 & $0.999$ & $0.109$ & $1.211e\text{-}5$ \\

            & Frame 3 & $0.999$ & $0.137$ & $3.405e\text{-}5$ \\
        \bottomrule
    \end{tabular}
    \caption{DetJac statistics across transformer model configurations and time-points.}
    \label{tab:detjac_summary}
\end{table}

\subsection{Evaluation of Sample Reliability via Downstream Tasks.} We first evaluate the utility of our synthesized longitudinal scans as a data augmentation strategy for downstream classification using the VGG3D network~\cite{rahman2025alzheimer} as the backbone. A key advantage of our framework is that it synthesizes deformations, which can be applied to both the input images and their ground-truth segmentation maps. This allows us to generate new samples of paired longitudinal image and segmentation labels. More specifically, we generate missing scans for each subject in the dataset and incorporate these synthetic images to augment the training set. We then evaluate the anatomical fidelity of the synthesized longitudinal scans by training a hippocampal segmentation model based on U-Net backbone~\cite{ronneberger2015u} on the OASIS dataset~\cite{lamontagne2019oasis}. Here, we use $53$ longitudinal sequences with the corresponding segmentation labels generated by FreeSurfer~\cite{billot2023synthseg} as ground truth.

\begin{figure*}[!htbp]
\begin{minipage}[t]{0.40\textwidth} 
    \includegraphics[width=\linewidth, trim={0.9cm 0.5cm 0.7cm 0.7cm}, clip]{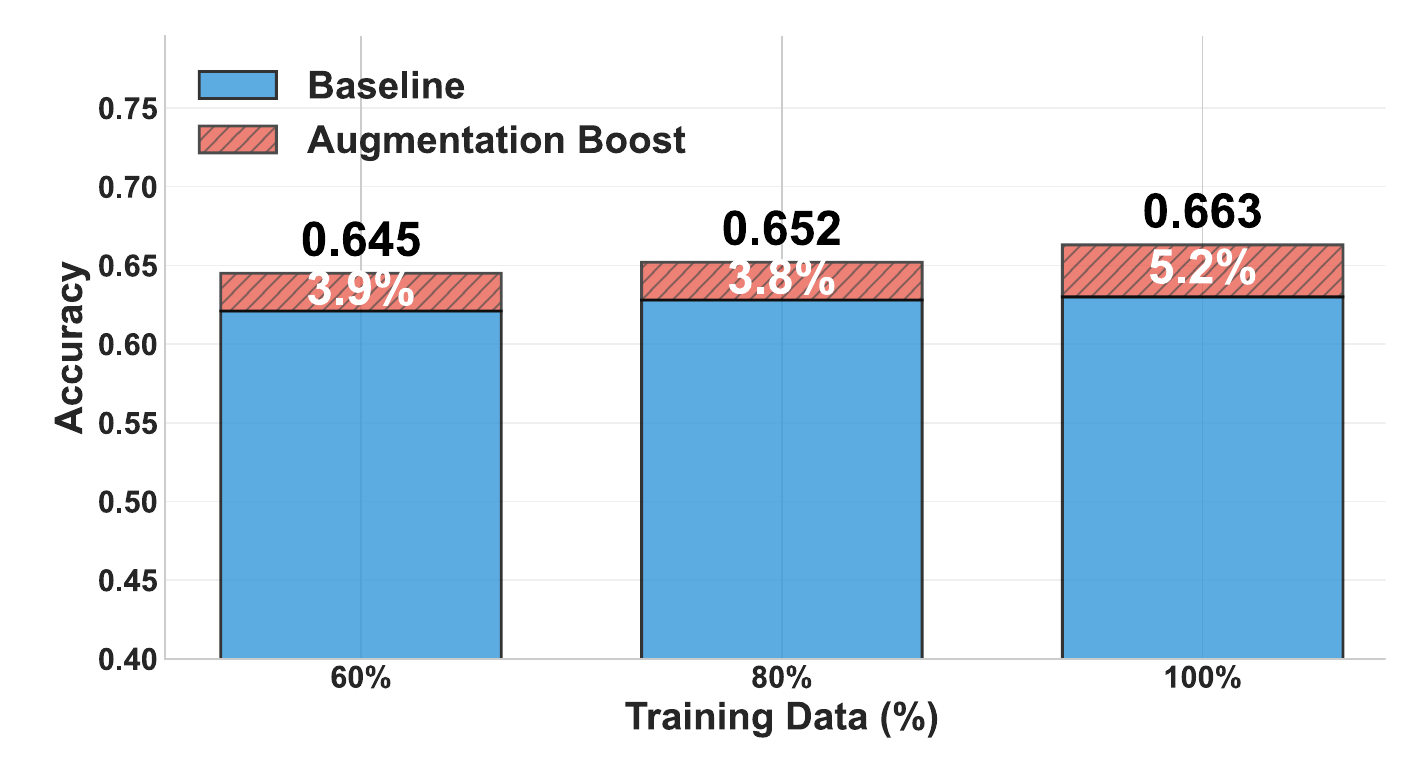}
    \caption{Improved ADNI classification accuracy via synthetic longitudinal MRI augmentation.}
    \label{fig:performance}
\end{minipage}
\hfill
\begin{minipage}[t]{0.55\textwidth}
    \includegraphics[width=\linewidth, trim={1cm 0.7cm 0.5cm 0.3cm}, clip]{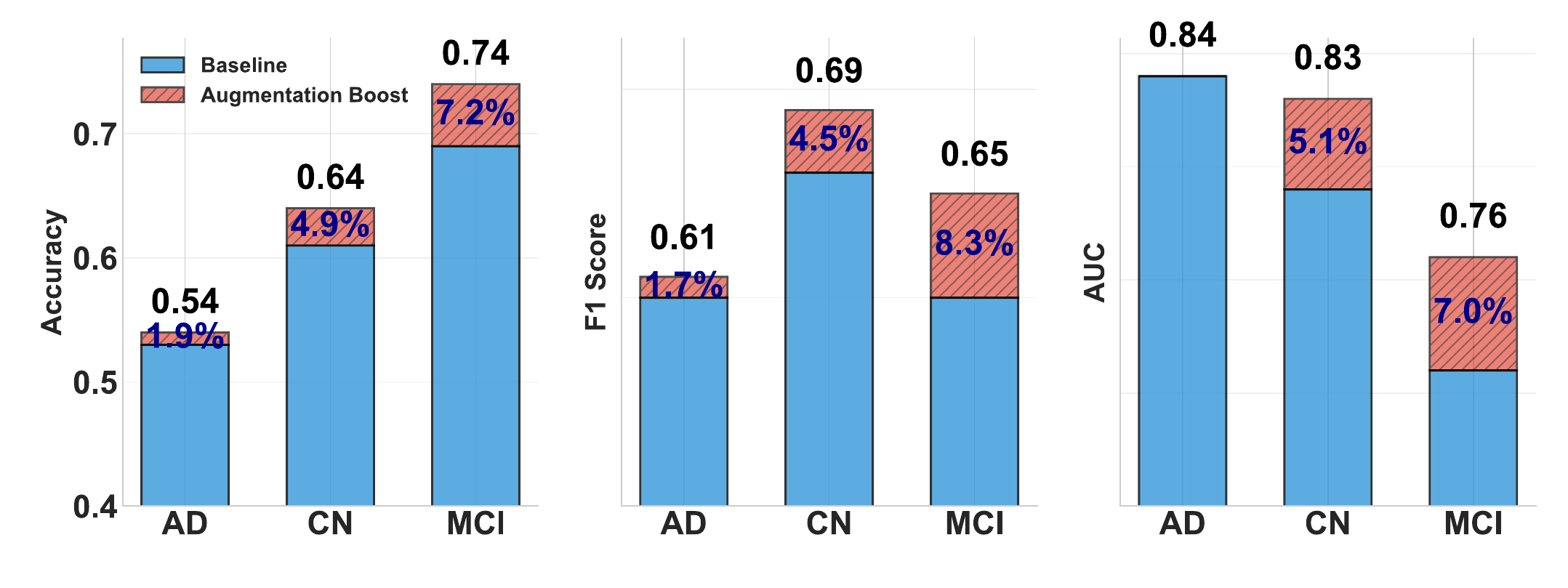}
    \caption{Per class performance improvement on ADNI with data augmentation using synthesized longitudinal MRI scans.}
    \label{fig:class_performance}
\end{minipage}
\end{figure*}

As shown in Figure~\ref{fig:performance}, we first gradually increase the proportion of original training data while keeping the synthesized sample fixed. The consistent improvements in classification accuracy across all training sizes demonstrate that our generated scans effectively supplement the real data, especially when labeled data is limited. Furthermore, Figure~\ref{fig:class_performance} presents class-wise improvements for CN, MCI, and AD, demonstrating that the synthesized scans boost performance across all disease stages.

\begin{figure*}[!ht]
\centering
\includegraphics[width=0.9\linewidth, height=0.27\textheight, trim={0.0cm 0.0cm 0cm 0cm}, clip]{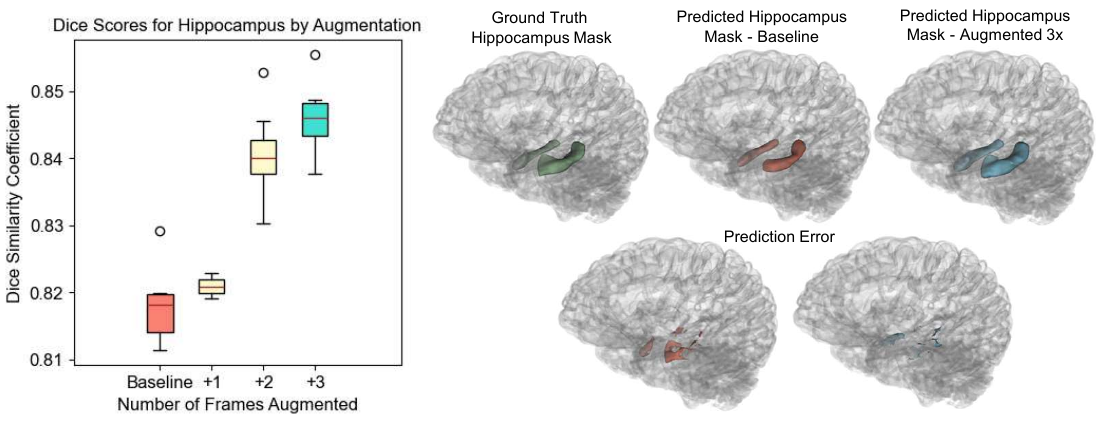}
\caption{Left: hippocampi segmentation results trained with frame-wise longitudinal augmentation from OASIS. Right (top): exemplary visualization of groundtruth hippocampi segmentations vs. predictions without/with augmentation. Right (bottom): segmentation error maps between groundtruth vs.predictions without/with augmentation.}
\label{fig:seg_results}
\end{figure*}

We then train the segmentation model with various numbers of synthesized longitudinal frames added to the original training set to further evaluate their potential impact on downstream performance. As shown in Figure~\ref{fig:seg_results} (Left), the inclusion of 1-3 synthetic volumes progressively improves the segmentation dice score (a metric that quantifies the overlap between the predicted segmentation and the ground truth)~\cite{dice1945measures}, yielding up to a 3\% increase over the baseline. Figure~\ref{fig:seg_results} (Right) presents qualitative comparisons: the baseline model under-segments the hippocampal tail and produces overly smoothed boundaries, while the augmented model generates segmentation masks that better align with the ground truth, more accurately capturing the full hippocampal anatomy with sharper contours. These results highlight the anatomical consistency of our generated scans and their utility in enhancing downstream segmentation performance. 

\subsection{Ablation Studies}
\paragraph*{\bf Transformer scalability} To evaluate the scalability of our proposed model in the deformation space (4DLoG-Def.), we experiment with three transformer variants, LDT-S, LDT-L, and LDT-XL, which differ in hidden embedding dimensions, depth and the number of attention heads. 

Table~\ref{tab:scaling_transformer} summarizes the corresponding model configurations (hidden dimension, \#attention heads, \#layers) and performance metrics. As the model capacity increases, we observe improvement in both the visual fidelity and the quantitative performance of the generated scans. This suggests that our architecture benefits from scaling, effectively using increased quality of learned representations to model temporally complex anatomical changes. The results further show that larger transformer backbones can more accurately capture longitudinal brain dynamics while maintaining structural coherence.
\begin{table}[!h]
    \centering
    \footnotesize
    \setlength{\tabcolsep}{6pt}
    \begin{tabular}{l c c c c c}
        \toprule
        Model & FID $\downarrow$ & PSNR $\uparrow$ & SSIM $\uparrow$ & KID $\downarrow$ & Configuration\\
        \midrule
        LDT-S  & $0.499 \pm$ \scriptsize{0.17} & $25.434 \pm$ \scriptsize{1.83} & $0.924 \pm$ \scriptsize{0.02} & $0.033 \pm$ \scriptsize{0.04} & (384, 6, 12)\\
        LDT-L  & $0.241 \pm$ \scriptsize{0.2} & $\mathbf{25.897 \pm}$ \scriptsize{1.30} & $\mathbf{0.934 \pm}$ \scriptsize{0.01} & $0.003 \pm$ \scriptsize{0.005} & (768, 12, 12) \\
        LDT-XL & $\mathbf{0.160 \pm}$ \scriptsize{0.09} & $25.379 \pm$ \scriptsize{1.35} & $0.927 \pm$ \scriptsize{0.005} & $\mathbf{0.002 \pm}$ \scriptsize{0.001} & (960, 12, 16) \\
        \bottomrule
    \end{tabular}
    \caption{Model architecture scaling performance metrics for different transformer configurations written as (Hidden Dimension, Number of Heads, Number of Layers).}
    \label{tab:scaling_transformer}
\end{table}

\paragraph*{\bf APPE's Age-wise Temporal Position Encoding} We provide qualitative results comparing two age-conditioning strategies: age-wise temporal position encoding applied to extracted volumetric patches, and linear age vectors incorporated through adaptive normalization alongside diffusion timestep $\tau$, disease class $y$, and anatomical prior $\nabla I_{a_0}$. 

\begin{figure*}[!ht]
\centering
 \includegraphics[trim={0.0cm 0.0cm 0cm 0cm}, width = 0.95\textwidth, clip]{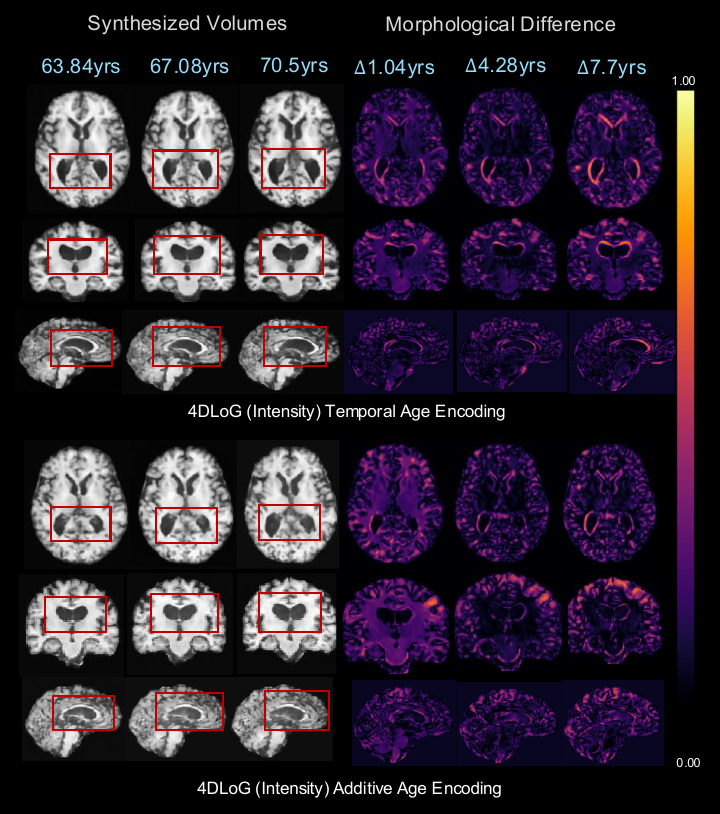}
\caption{Visualization of axial, coronal and sagittal views of scans synthesized by our intensity framework with and without age-specific temporal position encoding. Scans are generated for a subject with Alzheimer's Disease.}
\label{fig:temp_additive}
\end{figure*}

Figure~\ref{fig:temp_additive} indicates that temporal position encoding leads to better integration of age information, resulting in more realistic and anatomically plausible brain changes over time as compared to the linear additive approach.

\paragraph*{\bf Intensity vs. Deformation Variants} We implement an intensity-based version (4DLoG-Int.) by replacing the registration network with a latent-space VAE-GAN from MONAI~\cite{cardoso2022monai, larsen2016autoencoding}, trained on our dataset to ensure accurate reconstruction. We present results from both the intensity-based model and the deformation-based model, which learns the distribution of velocity fields from a pre-trained registration network. 

\begin{figure*}[!h]
\centering
 \includegraphics[trim={0.0cm 0.0cm 0cm 0cm}, width=1\textwidth, clip]{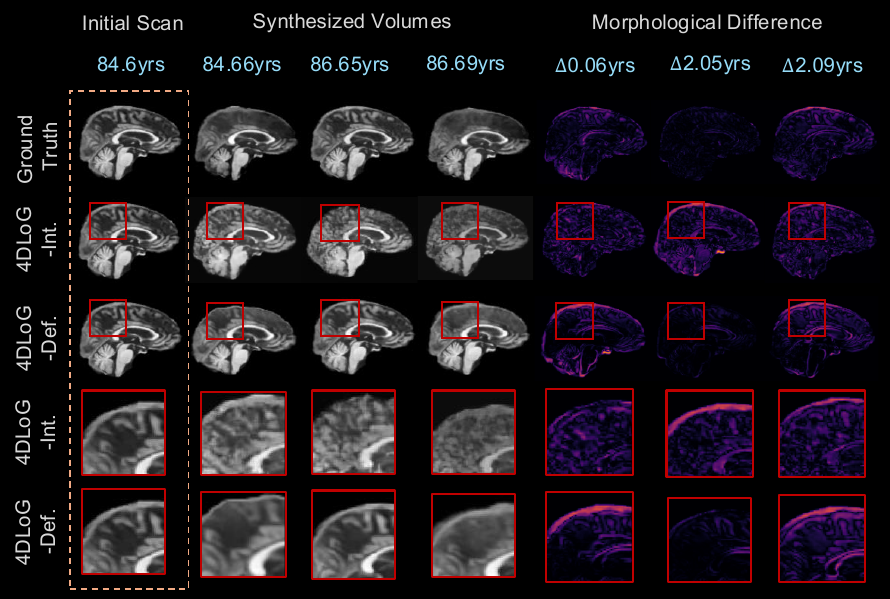}
\caption{Visualization of sagittal views of scans synthesized by our intensity framework vs. our deformation framework for a subject with Alzheimer's Disease.}
\label{fig:hallucinations}
\end{figure*}

Figure~\ref{fig:hallucinations} shows that the intensity model could produce unrealistic, hallucinated anatomical structures due to direct manipulation of voxel intensities. In contrast, the deformation model generates structure-preserving transformations by learning plausible velocity fields that deform the initial scan. This leads to more anatomically consistent and clinically relevant samples, making the deformation-based approach better suited for downstream tasks.

\subsection{Limitations and Future Directions}
\textcolor{black}{While our proposed model 4DLoG demonstrates promising results for longitudinal brain image generation and downstream applications, several limitations highlight potential opportunities for future research.}

\textcolor{black}{First, our two-stage design implies that registration error could influence the generative diffusion process. More specifically, an underestimated velocity field may leave genuine anatomical changes insufficiently accounted for by the spatial transformation, causing the generative model to absorb part of the geometric change into the appearance transformation. In contrast, an overestimated velocity field may introduce artificial geometric changes into the conditioning signal, potentially leading to corresponding artifacts in the generated image. This is particularly relevant for longitudinal cases in which a subject transitions from a healthy to a diseased stage, as the deformation distribution may differ from those observed in healthy-to-healthy or diseased-to-diseased pairs. Potential solutions could be including carefully incorporating age and disease-related information into a conditional registration prior,  or jointly optimizing registration and generation objectives to refine the estimated deformation fields based on their downstream impact on image generation. While this is beyond the scope of the current work, we consider it an important direction for future research.}

\textcolor{black}{Second, similar to the setting adopted by the existing methods~\cite{puglisi2024enhancing, pombo2023equitable} that do not assume future disease status is known at inference, our model does not explicitly infer the future disease status. In the current setting, the proposed 4DLoG generates the future scan based on the observed initial scan, as well as baseline disease diagnosis and demographic variables. A potential extension would be to jointly learn an auxiliary disease-transition prediction task, where an additional module predicts whether a subject is likely to remain healthy or transition to a diseased state. The predicted disease trajectory could then provide an additional conditioning signal to the generative model. However, effectively learning such a task would require a sufficiently large number of longitudinal cases with observed health-status transitions, which is currently limited in our available datasets. We therefore do not incorporate this additional task into the current model design.}

\textcolor{black}{Finally, we evaluate 4DLoG on both ADNI and OASIS, with OASIS exclusively used as an independent out-of-distribution testing cohort for longitudinal generation and downstream segmentation, while training and validating the model on ADNI. This demonstrates the out-of-distribution generalization of our method. Nevertheless, evaluation on additional longitudinal datasets would further strengthen the assessment of generalizability. In this work, while our group currently only has access to ADNI and OASIS, we plan to extend our evaluation to additional longitudinal brain MRI cohorts as they become available.}

\section{Conclusion}
In this work, we introduce 4DLoG, a novel framework for synthesizing a complete 4D longitudinal brain anatomy from a single baseline scan using a transformer-based diffusion model that operates in the space of diffeomorphic velocity fields. \textcolor{black}{In contrast to prior approaches that synthesize individual volumes pairwise or recurrently, our model performs a full 4D (3D$\times$T) diffusion process that jointly learns spatial and temporal dependencies across longitudinal sequences through newly designed spatio-temporal patch extraction and attention. Moreover, it explicitly models the distribution of topology-preserving deformations rather than voxel intensities, ensuring anatomically consistent trajectories across time.} 
\textcolor{black}{It is worth noting that, in contrast to 4D diffusion models in the natural video and computer vision space, we are, to the best of our knowledge, the first to model 4D longitudinal neuroimaging sequences in the space of smooth deformations. By explicitly modeling anatomical shape changes, 4DLoG can capture longitudinal structural dynamics while naturally accommodating the irregular and non-uniform temporal sampling characteristic of real-world longitudinal neuroimaging data.} Extensive experiments demonstrate the superiority of our approach over the state-of-the-art in terms of synthesis quality and downstream clinical utility, including neurodegenerative disease classification and brain segmentation. Notably, our flexible network architecture supports both intensity- and deformation-space modeling. By filling in gaps in sparse longitudinal datasets and predicting brain changes over time, our method represents a significant step toward data-driven modeling of neurodegenerative progression in clinical and research settings. Future work will unify intensity-based and deformation-based models to better capture longitudinal brain changes involving both structural deformations and appearance variations. \\

\noindent {\bf Acknowledgments.} This work was supported by NSF CAREER Grant 2239977 and NIH 1R21EB032597.\\

\noindent\textbf{Declaration of generative AI and AI-assisted technologies in the manuscript preparation process.} During the preparation of this work the author(s) used Copilot in order to perform a grammar check and make the content concise. After using this tool/service, the author(s) reviewed and edited the content as needed and take(s) full responsibility for the content of the published article.

\clearpage
\onecolumn

\bibliographystyle{elsarticle-harv} 
\bibliography{main}

\end{document}